# Extraction of Protein Sequence Motif Information using PSO K-Means

R. Gowri*, R. Rathipriya**

**Abstract**

The main objective of the paper is to find the motif information.The functionalities of the proteins are ideally found from their motif information which is extracted using various techniques like clustering with k-means, hybrid k-means, self-organising maps, etc., in the literature. In this work protein sequence information is extracted using optimised k-means algorithm. The particle swarm optimisation technique is one of the frequently used optimisation method. In the current work the PSO k-means is used for motif information extraction. This paper also deals with the comparison between the motif information obtained from clusters and biclustersusing PSO k-means algorithm. The motif information acquired is based on the structure homogeneity of the protein sequence.

**Keywords:** Biclustering, Clustering, PSO K-Means, Protein Motifs

## 1. Introduction

Clustering is one of the known data mining techniques used to group similar kind of data elements. It is used to discover similar patterns from sea of data.The similarity among objects in the same cluster is greater than in different clusters. It is widely used in many research areas like bioinformatics, pattern recognition, data mining, statistics, image analysis and machine learning. As all these areas are dealing with the unclassified data, the clustering is well suited to these kinds of research areas. The clusters can be found based on various similarities among the data such as intra distance and inter distance of the clusters. The quality of clusters will be evaluated based on our objective.

Proteins are present in every cell of the organisms. They are involved virtually in almost all cell activities. They are responsible for the various metabolic activities, nutrition transportation, regulations, etc. The protein plays a vital role in cellular processes. The protein consists of twenty amino acids. They possess different characteristics. It is great challenge to the bioinformatics that to find which combination of proteins perform what type of activities. The functionality of proteins are discovered by various methods like sequence-motif based method, homology based methods, structure based methods and so on. the motifs can be extracted from the clusters that are generated by various computational techniques.

Biclustering is another data mining technique. It is also named as co-clustering or two way clustering. It generates biclusters of different sizes and characteristics. The process of grouping data based on both the samples (genes) and attributes (conditions).

The major difference between the clustering and biclustering are as follows.

1. Clustering applied to either rows or columns of the dataset, but biclustering is applied to both rows and columns.
2. The size of any one of the dimensions of all the clusters will be same, but biclusters are of different size.
3. Biclustering groups more similar element than the clustering process.

___________________
* Research Scholar, Department of Computer Science, Periyar University, Salem, Tamilnadu, India.
  Email: gowri.candy@gmail.com
* Assistant Professor, Department of Computer Science, Periyar University, Salem, Tamilnadu, India.
  Email: rathi_priyar@yahoo.co.in

In our current work, the motifs are extracted from the clusters and biclusters that are detected using the PSO k-means approach. The quality of motifs extracted from clusters and biclusters are compared.

The organisation of the paper is as follows. The first section deals with the introduction to various concepts involved in this work. The second section summarizes themethods that are applied to the protein sequence in this work. The third section summarizes the preprocessing of protein sequence in order to apply the proposed model and the measures that are used to evaluate the quality of the clusters and biclusters. The fourth section shows the result of our work and brief description on them. The fifth section concludes the current work with future enhancement.

## 2. Methods and Material

The protein motifs are extracted using various data mining techniques. In this study we use the clustering technique. The k-means algorithm is used for this process. The PSO k-means algorithm is the proposed algorithm.

### 2.1. K-means Algorithm

The k-means algorithm is one of the simplest, ever using and known clustering algorithms. It is suitable for many research areas. It is numerical, unsupervised, non-deterministic and iterative clustering method. The given dataset is partitioned into K clusters. The data points are assigned randomly to the clusters resulting in clusters that have roughly the same number of data points. This algorithm works faster for the larger dataset. It produces the tighter clusters than hierarchical clustering. The simple k-means algorithm is shown in Figure 1.

**Figure 1:** Simple K-Means Algorithm

1. Select K objects from the space represented by the objects that are being clustered. These are initial group centroids.
2. Assign each object to the group that has the closest centroid.
3. When all objects have been assigned, recalculate the positions of the K centroids.
4. Repeat Steps 2 and 3 until the centroids no longer move. This produces a separation of the objects into groups from which the metric to be minimized can be calculated.

### 2.2. Particle Swarm Optimisation

Swarm intelligence is based on the collective behaviour of the decentralised and self-organised system of the natural swarms. The particle swarm optimisation is one of the swarm intelligence techniques. It is a stochastic optimisation technique. It is a population based method that is designed based on the birds flocking behavior. Selecting the initial population is decided based on the problem criteria. The number of individuals in the population is decided based on the objective of the problem. The particle at each time interval updates its position and velocity based on its own best position achieved so far and the global best position achieved by any one of the particles in its swarm. The random terms are used to weigh the acceleration of the particle. The best position is evaluated based on the objective function of the problem. The basic particle swarm optimisation algorithm is shown in Figure 2.

**Figure 2:** Particle Swarm Optimisation Algorithm

Generate the initial population by setting number of individuals in the population.

For each particle initialize the parameters like velocity and position.

While the iteration <= iteration max

Evaluate the fitness of each particle position and Update the 'pbest' and 'gbest' positions.

Update the particle position and velocity

end

### 2.3. Clustering using PSO K-Means

The k-means algorithm is one of the simplest, ever using and known clustering algorithms. K-means algorithm highly depends on the initial state and converges to local optimum solution. So the optimisation method is used to obtain the optimum result of the k-means algorithm. PSO is ideal for identifying the global optimal cluster i.e. global significant amino acid. This method suffers from local minima problem which can be eliminated by the optimisation technique like PSO. The number of particles is set as per the user's wish not greater than 100. It is set in order to overcome the complexity of the computation. The initial velocities for each particle are set to random values. The position of each particle is the cluster centroid. The number of clusters taken is fixed for each particle. The cluster intra distance is taken as the fitness in the PSO algorithm (Fitnessclus) using equation (1).

$$\text{Fitness}_{clus} = \frac{\sum_{k=1}^{nclust} \sum_{i=1}^{n} dist(A_k\ i, cent_k)}{nclust} \quad (1)$$

The particle with minimum fitness value is taken as the best particle. The various steps involved in PSO k-means are shown in Figure 3.

**Figure 3: Steps in PSO-k-Means Model**

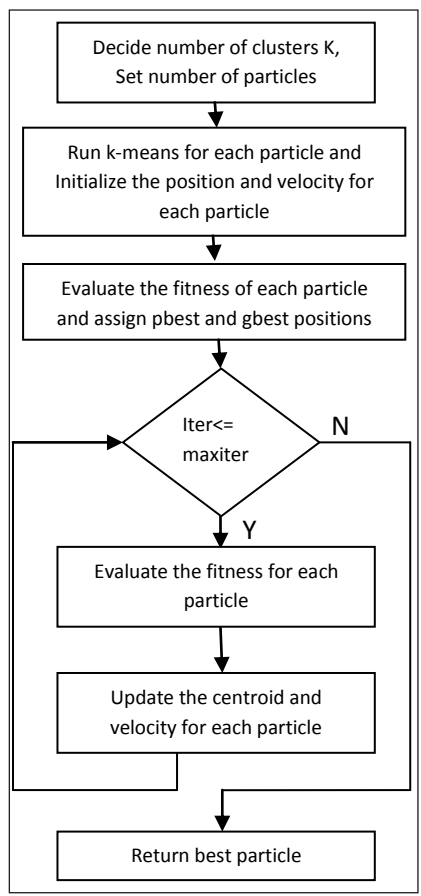

### 2.4. Biclustering Using PSO K-Means

Biclustering is viewed as an optimisation problem with the objective of finding the biclusters with minimum MSR and high volume. PSO is ideal for identifying the global optimal bicluster i.e. global optimal sequence motif. This method suffers from local minima problem which can be eliminated by the optimisation technique like PSO. A Biclustering algorithm along with Particle Swarm Optimisation (PSO) technique for protein sequence is proposed. The main objective of this algorithm is to identify the minimal subset of amino acids with coherent in their presence in the sequences. These biclusters have low MSR value, where MSR value is the quality measure for correlation between various residues and amino acids. PSO is initialised with initial biclusters which is obtained by using PSO K-Means on both the dimensions of the normalised protein frequency matrix A. This results in fast convergence compared to random particle initialisation of the PSO and it also maintains high diversity in the population. Each particle of PSO explores a possible solution. It adjusts its flight according to its own and its companions flying experience. The personal best (pbest) position is the best solution found by the particle during the course of flight. This is denoted by the pbest and the optimal solution is attained by the global best (gbest). BPSO updates iteratively the velocity of each particle towards the pbest and gbest. For finding an optimal or near optimal solution to the problem, PSO keeps updating the current generation of particles. This process is repeated until the stopping criterion is met or maximum number of iteration is reached. The MSR is taken as the fitness of the bicluster (Fitness$_{bic}$) using the equation (2).

$$\text{Fitness}_{bic} = \frac{1}{|I||J|} \sum_{i \in I, j \in J}(b_{ij} - b_{iJ} - b_{Ij} + b_{IJ})^2 \quad (2)$$

wherefitness is the mean square residue of a bicluster b with I rows and J columns, $b_{iJ}$ represents the column average and $b_{Ij}$ represents the row average.

## 3. Experimental Setup

### 3.1. Protein Sequence Dataset

In this study, more than 2000 protein sequences are extracted from the Protein Sequence Culling Server (PISCES) as the dataset. Out of this 300 samples were taken. In this database, none of the sequence shares more than 25% sequence identity. Sliding windows with nine consecutive residues are taken. Each window contains one sequence segment of nine continuous positions. In addition, all the sequences are considered homologous in the sequence database. Secondary structures are also taken from DSSP. DSSP is a database of secondary structure assignments for all protein entries in the PDB.

### 3.2. Sequence Representation

Each sequence is represented using sliding window. A sliding window containsre sidues as rows and amino acids as columns. In this study nine consecutive residues are taken. There are twenty amino acids are present in protein. The size of each sequence differs from the other. In order to make all the sequence of same size and to convert the categorical data into numerical the following steps are performed. First arrange the one-dimensional sequence data into two-dimensional by placing every nine consecutive residues as one column. The number of rows will be same but the column size will differ for all the sequences. Then fill the empty spaces in last column with any constant value. Now find the frequency of each amino acid in each row and generate the frequency window for every sequence. Now each sequence will be in the size 9 X 20. The secondary structures of the protein sequence are collected from the DSSP database. There are eight classes in secondary structure which is mapped to three classes the following conversion model: assigning H, G, and I to H (Helices), assigning B and E to E (Sheets), and assigning all others to C (Coils).

These sliding windows are used for clustering process, whereas for biclustering the each window is normalised to a single row. For this purpose the normalisation techniques like average value, difference between max and min value and more frequent value are taken. Now the data become two-dimensional with each row representing a sequence.

### 3.3. Distance Measures

The city block distance is suitable for this type of data. Since this measure deals with n-dimensional data element it is suitable for this type of sequence profiles. The distance measure is calculated using the equation (3).

$$\text{Dissimilarity} = \sum_{i=1}^{R} \sum_{j=1}^{A} |V_k(i,j) - V_c(i,j)| \quad (3)$$

where R is the number of residues and A represents 20 different amino acids. The $V_k(i,j)$ is the value of the frequency matrix at row i and column j of $k^{th}$ sequence. The $V_c(i,j)$ is the value of the frequency matrix at row i and column j of that represents the centroid of the cluster.

### 3.4. Cluster Validity Measure

The clusters detected by each particle are evaluated for their fitness. The intra-cluster distance is used as the cluster validity measure. The intra-distance of the cluster shows how close the elements of the cluster are placed. The average intra distance value of each particle is calculated. The particle with the minimum distance is taken as the best particle.

### 3.5. Biclustervalidity Measure

The biclusters detected by each particle are evaluated for their fitness. The Mean Square Residue (MSR) value is used to validate the biclusters of each particle. The particle with minimal MSR value is considered as best particle.

$$M(I,J) = \frac{1}{|I||J|} \sum_{i \in I, j \in J}(b_{ij} - b_{iJ} - b_{Ij} + b_{IJ})^2 \quad (4)$$

whereM (I, J) is the mean square residue of a bicluster b with I rows and J columns, $b_{iJ}$ represents the column average and $b_{Ij}$ represents the row average.

### 3.6. Structure Similarity Measure

Cluster's average structure is calculated using the following formula:

$$\frac{\sum_{i=1}^{ws} \max(P_{i,H}, P_{i,E}, P_{i,C})}{ws} \quad (5)$$

where ws is the window size and $P_{i,H}$ shows the frequency of occurrence of helix among the segments for the cluster in position i. $P_{i,E}$ and $P_{i,C}$ are defined in a similar way. If the structural homology for a cluster exceeds 70%, the cluster can be considered structurally identical. If the structural homology for the cluster is between 60% and 70%, then cluster can be considered weakly structurally homologous.

## 4. Results and Discussions

The results of both the clustering and biclustering process using PSO kmeans model as proposed in our work are tabulated and interpreted below. The comparison between the clustering and biclustering process is performed based on their secondary structure similarity is discussed. The significant amino acids of the protein sequence are extracted and visualized using logo representation. The motifs detected for the given sequences are taken as superset for the significant amino acids and their interpretation is also tabulated.

### 4.1. Structural Similarity Comparison

The clustering technique is quite inefficient when compared to the biclustering process. This is proven from the secondary structure similarity value. The number of clusters with similarity generated by our model is less when compared to the number of biclusters generated by the same. Table 1 shows that biclustering can detect the Biclusters with more similarity value. So the biclustering technique is better when compared to the clustering technique for protein motif extraction process. The structure similarity value is calculated using the equation. The clusters or biclusters with similarity value greater than 70% is taken as highly homologous and with similarity greater than 60% and less than 70% is taken as weakly homologous sequence.

**Table 1:** Comparison between Clustering and Biclustering Based on Secondary Structure Similarity

| Structure similarity value >= | Clusters | Biclusters |
|---|---|---|
| 70% | 1 | 3 |
| 65% | 3 | 5 |
| 60% | 4 | 5 |

**Figure 4:** Cluster 1 Amino Acid Logo Representation

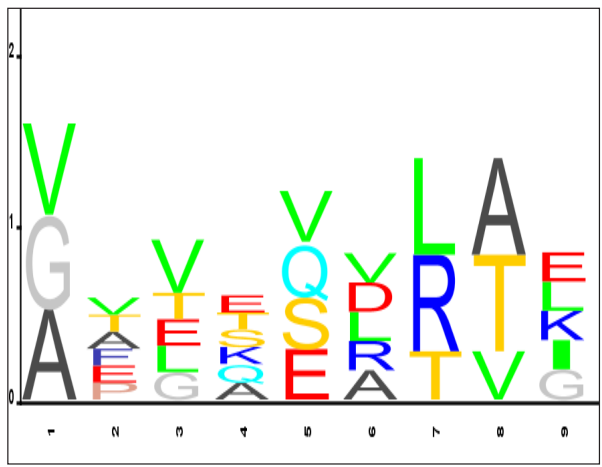

### 4.2. Sequence Motif Representation

The PSO k-means-Biclustering Model is chosen to improve the cluster quality which extracts the motif information efficiently without converging to the local minima. The corresponding amino acid sequence columns that are purged by the PSO k-means Biclustering Model (Motif) are shown to be either the full or the partial superset of the significant amino acids (SAA): amino acids with appearing frequency greater than 7%. Table 5 and 6 illustrate the SAA and Motif relationship for the entire nine positions of the cluster 1 and 2. Thus, bi-clustering is proven to have the potential to select meaningful amino acids that biologists are interested at. The horizontal axis shows the positions and the vertical axis shows the bits of amino acids (Figure 4 and 5).

**Figure 5:** Cluster 2 Amino Acid Logo Representation

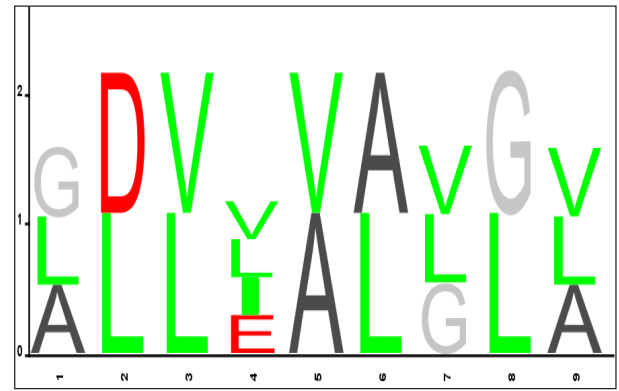

**Table 2:** Full and Partial Superset of PROTEIN SEQUENCE

MOTIF on Cluster 1

| Position | SAA | Motif | SAAMotif |
|---|---|---|---|
| 1 | AGV | ADEGILKTV | Full |
| 2 | AEFPTV | ADEGILKTV | Partial |
| 3 | EGLTV | ADEGILKTV | Full |
| 4 | AEKQS | ADEGILKTV | Partial |
| 5 | EQSV | RNQFPSY | Partial |
| 6 | ADLRV | ADEGILKTV | Partial |
| 7 | LRT | ADEGILKTV | Full |
| 8 | ATV | ADEGILKTV | Full |
| 9 | EGIKL | ADEGILKTV | Full |

**Table 3:** Full and Partial Superset of PROTEIN SEQUENCE

MOTIF on Cluster 2

| Position | SAA | Motif | SAAMotif |
|---|---|---|---|
| 1 | AGL | ADEGILKTV | Full |
| 2 | DL | ADEGILKTV | Full |
| 3 | LV | ADEGILKTV | Full |
| 4 | EILV | ADEGILKTV | Full |
| 5 | AV | ADEGILKTV | Full |
| 6 | AL | ADEGILKTV | Full |
| 7 | GLV | ADEGILKTV | Full |
| 8 | GL | ADEGILKTV | Full |
| 9 | ALV | ADEGILKTV | Full |

The PSO k-means-Biclustering Model is chosen to improve the cluster quality because it removes the columns of the data matrix in addition to eliminating the rows. The corresponding amino acid sequence columns that are purged by this Biclustering Model (Motif) are shown to be either the full orthe partial superset of the significant amino acids (SAA): amino acids with appearing frequency greater than 7%. Table 5 and 6 illustrate the SAA and Motif relationship for the entire nine positions of the cluster 1 and 2. Thus, bi-clustering is proven to have the potential to select meaningful amino acids that biologists are interested at.

## 5. Conclusion

The motif information is extracted using this PSO-k-means model efficiently than simple k-means. The results show that the PSO-k-means Biclustering Model is capable of preserving more original data points and obtaining better cluster quality. The motifs extracted are helpful for biologist for their further research works on new protein sequence. This will be an apt computational process for the biologists handling the two dimensional sequences. It reduces the computational cost when compared to the computations on multi-dimensional sequence analysis. The future enhancements will be performed in choosing number of residues, selecting number of clusters by using any criteria, predicting biological functionality of the extracted motif.